\documentclass[sn-mathphys-num]{sn-jnl}

\usepackage{graphicx}%
\usepackage{multirow}%
\usepackage{amsmath,amssymb,amsfonts}%
\usepackage{amsthm}%
\usepackage{mathrsfs}%
\usepackage[title]{appendix}%
\usepackage{xcolor}%
\usepackage{textcomp}%
\usepackage{manyfoot}%
\usepackage{booktabs}%
\usepackage{algorithm}%
\usepackage{algorithmicx}%
\usepackage{algpseudocode}%
\usepackage{listings}%
\usepackage{soul}

\usepackage{multirow}
\usepackage{subfig}
\usepackage{tikz,siunitx}
\usetikzlibrary{shapes.geometric,shapes.symbols}
\newcommand{\tikzsymbol}[2][circle]{\tikz[baseline=-0.5ex]\node[inner sep=2pt,shape=#1,draw,#2]{};}%
\definecolor{cadmiumgreen}{rgb}{0.0, 0.42, 0.24}
\definecolor{mediumcandyapplered}{rgb}{0.89, 0.02, 0.17}
\definecolor{cottoncandy}{rgb}{1.0, 0.74, 0.85}
\definecolor{lightpink}{rgb}{1.0, 0.71, 0.76}
\definecolor{pastelpink}{rgb}{1.0, 0.82, 0.86}

\renewcommand\hl[1]{#1}  

\theoremstyle{thmstyleone}%
\theoremstyle{thmstyletwo}%

\theoremstyle{thmstylethree}%

\begin{document}

\title[Article Title]{A Robust Image Forensic Framework Utilizing Multi-Colorspace Enriched Vision Transformer for Distinguishing Natural and Computer-Generated Images}

\author[1]{\fnm{Manjary} \sur{P. Gangan}}\email{manjaryp\_dcs@uoc.ac.in}

\author*[2]{\fnm{Anoop} \sur{Kadan}}\email{a.kadan@soton.ac.uk}

\author[1]{\fnm{Lajish} \sur{V. L.}}\email{lajish@uoc.ac.in}

\affil[1]{\orgdiv{Department of Computer Science}, \orgname{University of Calicut}, 
\country{India}}

\affil[2]{\orgdiv{School of Psychology}, 
\orgname{University of Southampton}, 
\country{United Kingdom}}



\abstract{\hl{The digital image forensics based research works in literature classifying natural and computer generated images primarily focuses on binary tasks. These tasks typically involve the classification of \textit{natural images versus computer graphics images} only or \textit{natural images versus GAN generated images} only, but not natural images versus both types of generated images simultaneously. Furthermore, despite the support of advanced convolutional neural networks and transformer based architectures that can achieve impressive classification accuracies for this forensic classification task of distinguishing natural and computer generated images, these models are seen to fail over the images that have undergone post-processing operations intended to deceive forensic algorithms, such as JPEG compression, Gaussian noise addition, etc. In this digital image forensic based work to distinguish between natural and computer-generated images encompassing both computer graphics and GAN generated images, we propose a robust forensic classifier framework leveraging enriched vision transformers.} By employing a fusion approach for the networks operating in RGB and YCbCr color spaces,  we achieve higher classification accuracy and \hl{robustness against the post-processing operations of JPEG compression and addition of Gaussian noise.} Our approach outperforms baselines, demonstrating 94.25\% test accuracy with significant performance gains in individual class accuracies. Visualizations of feature representations and attention maps reveal improved separability as well as improved information capture relevant to the forensic task. This work advances the state-of-the-art in image forensics by providing a generalized and resilient solution to distinguish between natural and generated images.}




\maketitle

\section{Introduction}
\label{sec_intro_mcevit}

\hl{The widespread availability of image acquiring devices has led to images being captured easily. Apart from the camera-captured \textit{`natural or real images'}, there exists \textit{`computer graphics images'}, i.e., images generated through computer graphics software. The other category of computer generated images includes the Generative Adversarial Network \textit{`(GAN) images'}. Beyond the artistic and entertaining purposes of these computer generated images, they also have a significant possibility to easily propagate misinformation via online platforms, especially when associated with fake news \mbox{\cite{anoop2019leveraging}}. Hence there exists a high necessity for computational aids to distinguish natural and computer generated images.} Most works in the literature that distinguish natural and computer generated images consider either computer graphics images only or GAN images only in the category of computer generated images \cite{tokuda2013computer,quan2018distinguishing,nataraj2019detecting,meena2021distinguishing}. That is, these works focus only on one category of computer generated images and not both. However, these forensic algorithms do not suit the real-world scenario where the process of image generation is unknown. Also, there could be cases if a generated image is not computer graphics then maybe it could be a GAN image or vice-versa. Hence a single forensic classification system is required to investigate images and understand whether they are natural/real photographs or the ones created by computer graphics or GAN algorithms, with high accuracy. 

With the advent of convolutional neural network (CNN) architectures and transformer based architectures, the image classification systems are achieving high classification accuracies \hl{\mbox{\cite{zhao2024review,liu2023survey}}}. Utilizing these techniques the forensic algorithms or tools to distinguish natural and computer generated images while moving to attain some success have been mostly vulnerable to different variations in images caused due to image quality, resolution, compression quality, or color which are capable enough to dramatically modify or restructure the underlying image properties\footnote{\url{https://www.nytimes.com/interactive/2023/06/28/technology/ai-detection-midjourney-stable-diffusion-dalle.html?auth=register-google&utm_source=pocket-newtab-intl-en}}. Image forgeries are almost always followed by these operations like adding some noise or applying JPEG compression to obscure any traces of image generation or forgeries and thereby deceiving image forensic algorithms and tools \hl{\mbox{\cite{wu2021towards}}}. Therefore, besides producing a high performance system for classifying photographs and computer generated images of both categories (computer graphics and GAN), it is equally essential to review the robustness of these forensic systems towards various post-processing operations. Hence, unlike the works in the literature that deals the forensic task of distinguishing natural and computer generated images as a binary-class classification task either by considering only computer graphics or only GAN images, our work proposes a generalized forensic algorithm, i.e., a three-class classification task classifying real, computer graphics and GAN images. Our methodology focuses on building a classification model such that it achieves high accuracy and is highly robust against post-processing operations. 
\\The major contributions of our work are:
\begin{itemize}
    \item \hl{We propose a robust image forensic classification framework for distinguishing natural and computer generated images including both computer graphics and GAN images, utilizing a vision transformer based approach and employing a fusion of two color space transformations RGB and YCbCr.}
    \item \hl{We devise a novel methodology by exploiting the vulnerability of the decrease in classification accuracies of image classes with the increase in compression factors, to majorly focus on building a forensic classifier system that is highly robust against the post-processing operations.}
    \item \hl{We compare the performance of our proposed model with a set of baselines and could observe that the performance of our approach outperforms the baselines with significant performance gains in total and individual class accuracies, is highly robust against post-processing operations of image compression and addition of gaussian noise, and is also generalizable to unseen test data.}
    \item \hl{We visualize the feature representations of our model and compare with the input image features and feature representations of the baseline method, where we could observe that our model provides much better separability between the different classes, helping towards improved classification performance.}
    \item \hl{We also analyze the attention maps of the networks of our fused model which shows that our novel methodology has the capability of improved information capture relevant to the forensic task of classifying natural and generated images.}
\end{itemize}

The rest of this paper is organized as follows. Section \ref{sec_related_mcevit} presents a brief survey of related works and delineates our proposed work from the related works in the literature. Section \ref{sec_methodology_mcevit} discusses the methodology of the proposed work along with the dataset used in this study, motivation and detailed description of the proposed model. Section \ref{sec_exp_mcevit} presents the experimental settings and details of the baseline models used for comparing our proposed model. Section \ref{sec_results_mcevit} presents the results and discussion including the results of the experiments for robustness, generalizability, feature visualization, and the analysis of attention maps. Section \ref{sec_conclusion_mcevit} finally draws the conclusions.

\section{Related work}
\label{sec_related_mcevit}

Image processing has a wide range of applicability in various research areas, including computer vision \cite{li2024soccer,zhao2024gan}, computer graphics \cite{sheng2018intrinsic}, medical image processing \cite{ali2024egdnet,dai2021deep,guan2023artificial}, digital image forensics \cite{diwan2024advancing,gangan2022distinguishing,diwan2024cnn,coccomini2022combining,diwan2023unveiling,gangan2023exploring}, etc. \hl{Digital Image Forensics is a research area that examines digital images using scientific approaches to provide evidence of their authenticity in different contexts, such as criminal cases, civil cases, etc. There are various different sub-tasks, which this area of research focuses on. These mainly include the development of computational algorithms to detect computer-generated images such as graphics and GAN images \mbox{\cite{tokuda2013computer,quan2018distinguishing,nataraj2019detecting,meena2021distinguishing}}, detecting recaptured images \mbox{\cite{hussain2022evaluation}}, forged images \mbox{\cite{mehrjardi2023survey}}, etc.}

Image forensic based works that distinguish natural and computer graphics images can be seen since the 1990s. \hl{These include statistical feature based works that utilize features based on local binary patterns, co-occurrence matrices, histogram features, etc. \mbox{\cite{wang2014statistical,peng2017discrimination}}, transform domain features based works utilizing wavelet features, quaternion wavelet features, etc. \mbox{\cite{ozparlak2011differentiating,wang2017forensics}}, device based features exploiting features based on intrinsic fingerprints, demosaicing, etc. \mbox{\cite{swaminathan2008digital,gallagher2008image}}. Also, there have been many works utilizing deep learning architectures such as CNNs \mbox{\cite{rahmouni2017distinguishing,de2018exposing,quan2018distinguishing,meena2021distinguishing}.} Works that distinguish natural and GAN images are comparatively much more recent than the category described above of natural versus computer graphics images. These works in the literature that attempt natural versus GAN images, mostly employ deep learning based approaches such as CNNs, XceptionNet, InceptionNet, etc. \mbox{\cite{nataraj2019detecting,marra2018detection,hsu2020deep,marra2019incremental}}, than the handcrafted feature based approaches \mbox{\cite{li2020identification}}.} Few works are also seen to utilize transformer based architectures to detect GAN generated videos \cite{coccomini2022combining}, but these works also only consider the single category of generation, that is natural versus GAN only. \hl{Works detecting GAN images generally consider image content based feature analysis \mbox{\cite{marra2018detection,karakose2024new,samal2023obscene}} or some works are also found to consider residual features \mbox{\cite{li2020identification}}, biometric features \mbox{\cite{pasquini2023identifying}}, etc. GAN image detection algorithms have applicability in numerous categories such as detection of generated faces \mbox{\cite{hsu2020deep,he2019detection,pasquini2023identifying}}, generated obscene images \mbox{\cite{samal2023obscene}}, generated medical images \mbox{\cite{karakose2024new}}, etc. Works in the literature classifying natural and computer-generated images are also found to study and conduct robustness experimentations towards post-processing operations, mainly JPEG compression \mbox{\cite{peng2017discrimination,marra2018detection,pasquini2023identifying}}. Table \mbox{\ref{table_literature_mcevit}} shows a snapshot of works classifying natural and computer-generated images.}

\begin{table}[!h]
\centering
\caption{Related works classifying natural and computer-generated images}
\label{table_literature_mcevit}
\begin{tabular}{lll}
\toprule
Work & Feature/Method & Category of generated image \\ \midrule
Quan et al. \cite{quan2018distinguishing} & Convolutional neural network & Computer graphics images \\
Meena et al. \cite{meena2021distinguishing} & Two-stream CNN & Computer graphics images \\
Wang et al. \cite{wang2014statistical} & Texture similarity feature & Computer graphics images \\
Peng et al. \cite{peng2017discrimination} & Histogram features & Computer graphics images \\
Ozparlak et al. \cite{ozparlak2011differentiating} & Contourlet based features & Computer graphics images \\
Wang et al. \cite{wang2017forensics} & Quaternion wavelet features & Computer graphics images \\
Swaminathan et al. \cite{swaminathan2008digital} & Intrinsic fingerprints & Computer graphics images \\
Gallagher et al. \cite{gallagher2008image} & Demosaicing & Computer graphics images \\
Rahmouni et al. \cite{rahmouni2017distinguishing} & Convolutional Neural Network & Computer graphics images \\
Rezende et al. \cite{de2018exposing} & ResNet-50 & Computer graphics images \\
Nataraj et al. \cite{nataraj2019detecting} & Co-occurrence matrix \& CNN & GAN images \\
Marra et al. \cite{marra2018detection} & XceptionNet & GAN images \\ 
Hsu et al. \cite{hsu2020deep} & Pairwise Learning & GAN images \\
Marra et al. \cite{marra2019incremental} & Incremental learning & GAN images \\
Li et al. \cite{li2020identification} & Color based features & GAN images \\
He et al. \cite{he2019detection} & Color features and CNN & GAN images \\ 
Karaköse et al. \cite{karakose2024new} & Yolo models & GAN images \\ 
Pasquini et al. \cite{pasquini2023identifying} & Biometric features & GAN images \\ 
Gangan et al. \cite{gangan2022distinguishing} & Color features and EfficientNet & Computer graphics \& GAN images \\
\botrule
\end{tabular}
\end{table}

\subsection{Our work in context}

To the best of our knowledge, there is only a single work that distinguishes natural versus computer generated images by considering both computer graphics and GAN images in the category of computer generated images \cite{gangan2022distinguishing}, and utilizes an EfficientNet CNN architecture for the three-class classification task. But, to the best of our knowledge, no works are seen to be reported using the high performance recent deep learning architectures i.e., transformer based architectures for classifying natural images and both categories of computer generated images, i.e., Graphics and GAN generated images. \hl{Our digital image forensic analysis based work utilizes vision transformers and proposes a fusion based approach to distinguish between natural versus computer generated images of both categories, i.e., Graphics and GAN images. Our methodology is based on the novel idea of exploiting the vulnerability of decline in classification accuracies of different classes with the increase in compression, that helps to primarily focus on developing a forensic classifier system that is highly robust against post-processing operations such as JPEG compression, addition of gaussian noise, etc.} Despite other works in the literature that utilize certain color space transformations for the task of distinguishing natural images from computer-generated images of either of the category such as \cite{he2019detection,li2020identification}, our work utilizes two color spaces in our methodology, based on our rigorous set of experiments, one for improving the overall accuracy of the task of distinguishing natural images from computer-generated images of both categories and other color space that deals with improving the robustness of the proposed model against post-processing operations. Our methodology follows a transfer learning approach which helps to improve the accuracy of the task without involving the burdensome process of pre-training and increasing the training complexity.

\section{Methodology}
\label{sec_methodology_mcevit}

In this work, the forensic task of distinguishing natural and computer generated images is formulated as a three-class classification task, where the three classes are GAN, Graphics, and Real. GAN and Graphics images, even though both being computer generated, are maintained as separate classes as they have different generation processes. For this three-class classification problem, we utilize transformer based deep neural networks to find the best-fit mapping function $M: y= M(x)$ for the train data ${(x_1,y_1),(x_2,y_2),...,(x_n,y_n)}$ where for each image $x_i$ in the training set, $y_i \in (0,1,2)$, indicating either of the classes GAN, Graphics or Real, respectively.

\subsection{Dataset}
\label{sec_dataset_mcevit}

In this work, we utilize the GAN, Graphics, and Real images from the dataset used in the study \cite{gangan2022distinguishing}. The dataset comprises 12000 images in total, where each of the classes contains 4000 images. The class GAN consists of images collected from a variety of generative algorithms such as ProgressiveGAN \cite{karras2017progressive}, StyleGAN \cite{karras2019style}, StyleGAN2 \cite{karras2020analyzing}, and StyleGAN2-ADA \cite{Karras2020ada}. Whereas, the images for the classes Graphics and Real are obtained from the Computer Graphics versus Photographs dataset \cite{tokuda2013computer}. The entire dataset is challenging and contains a diverse category of images in terms of image content (indoor and outdoor scenes, animals, objects, etc.), origin (different generative/graphics algorithms or cameras, etc), and quality. Also, the computer generated images i.e., both GAN and graphics are photorealistic, i.e., they are not manually easy to predict as computer generated ones. \hl{There are no any pre-processing operations performed on the images in the dataset.} For the training, validation, and testing the dataset is proportionally split in the ratio 60:20:20 respectively.

\subsection{Motivation}
\label{sec_motivation_mcevit}

\begin{figure}[!h]
\centering
\includegraphics[width=.7\textwidth,keepaspectratio]{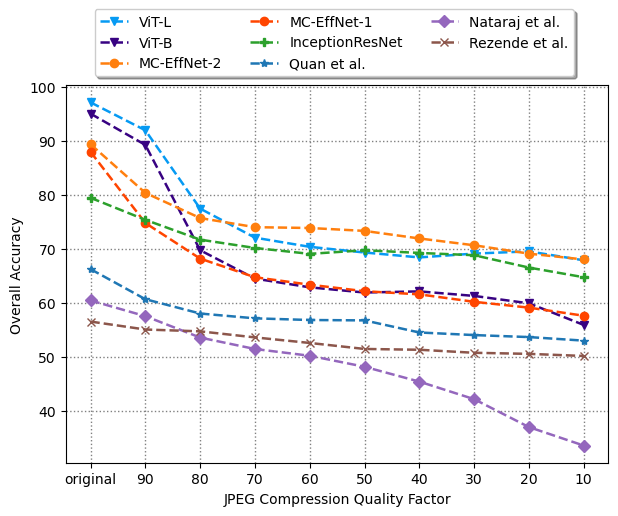}
\caption{\hl{Decline in classification accuracies of the models for the same set of images at varying levels of JPEG compression}}
\label{fig_compression_mcevit}
\end{figure}

The images transmitted through social media and online platforms are subject to image compression, mostly JPEG compression, where different online platforms follow different JPEG compression standards \cite{chuman2017image,mandelli2020training,wang2022jpeg}. Also in the cases of intentionally propagating fake images to support some piece of fake news to convince the audience, these images are seen to be mostly subject to multiple image compressions \cite{wu2021towards,anoop2019leveraging}. Hence, any model proposed for distinguishing natural and computer generated images, apart from obtaining high classification accuracy must also be highly robust towards compression. But most of the computational models in the literature proposed for the forensic task of distinguishing natural and computer generated images, are less robust to post-processing operations, particularly the JPEG compression \cite{barni2018adversarial}. That is, as the compression factor of the images increases (or the quality factor of the images decreases), the accuracy of the computational models to distinguish natural and computer generated images decreases. We analyze the classification accuracies of different computational models to distinguish GAN, graphics, and real images, at varying levels of JPEG compression, including the state-of-the-art vision transformer based architectures ViT-Base and ViT-Large \cite{dosovitskiy2021image}, state-of-the-art CNN based architecture InceptionResNet \cite{szegedy2017inception} and a few other baselines that classify natural and computer generated images \cite{gangan2022distinguishing,quan2018distinguishing,de2018exposing,nataraj2019detecting}. \hl{Figure \mbox{\ref{fig_compression_mcevit}} shows the classification accuracies of these models for original images (uncompressed images or zero compression factor)  and for various compression quality factors in descending order. As can be seen from the figure, the accuracy of all the models decreases from the original or uncompressed scenario, as the compression increases.} 

\begin{figure}[!h]
\centering
\subfloat[GAN class accuracies]
    {\includegraphics[width=.5\textwidth,keepaspectratio]{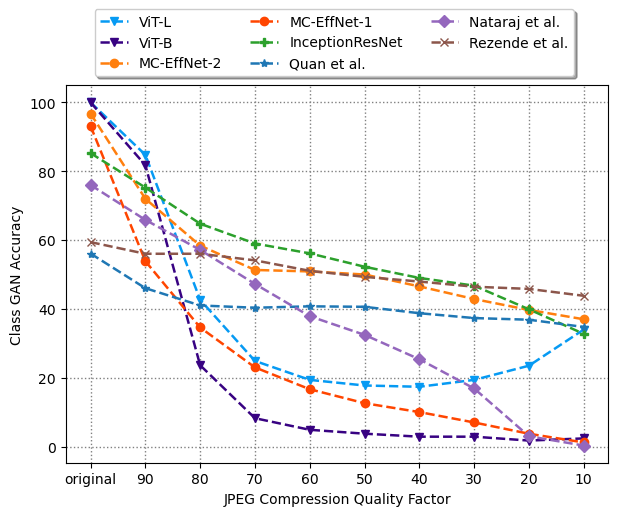}\label{fig_compression_gan_mcevit}}
\hfil
\subfloat[Graphics class accuracies]
    {\includegraphics[width=.5\textwidth,keepaspectratio]{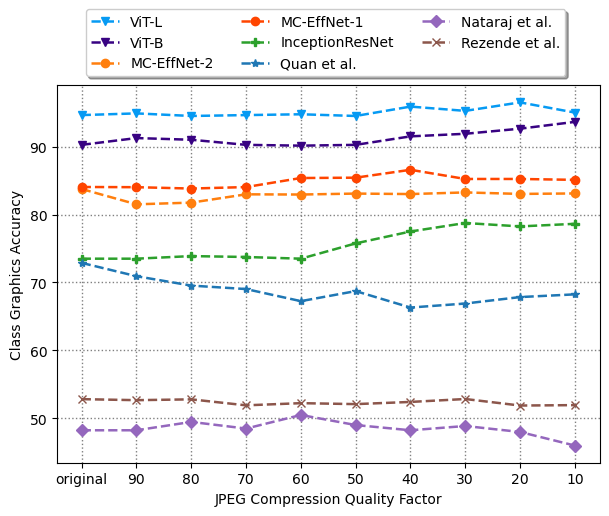}\label{fig_compression_graphics_mcevit}}
\hfil
\subfloat[Real class accuracies]
    {\includegraphics[width=.5\textwidth,keepaspectratio]{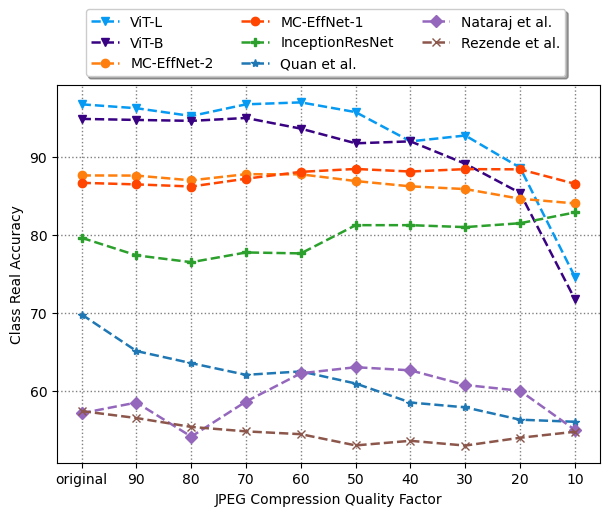}\label{fig_compression_real_mcevit}}
\caption{\hl{Rate of decrease in accuracies due to compression differs for different classes}}
\label{fig_compression_class_mcevit}
\end{figure}

To propose a robust model for the task, hence, we further analyze the class accuracies of these models. Figure \ref{fig_compression_class_mcevit} shows the individual class accuracies of the models for varying levels of JPEG compression i.e., accuracies of the class GAN generated images at varying levels of compression factors in figure \ref{fig_compression_gan_mcevit}, graphics generated images in figure \ref{fig_compression_graphics_mcevit}, and real images in figure \ref{fig_compression_real_mcevit}. We could observe that, for \hl{original or uncompressed images}, class accuracies of GAN generated images are always higher than the graphics and real images, for all the models. But at the same time, the class GAN is the most affected as the compression increases (as shown in figure \ref{fig_compression_gan_mcevit}). That is, there is a very high drop/decay in the class GAN accuracies of the models with even small increments in JPEG compression. For the class Real in figure \ref{fig_compression_real_mcevit}, the drop in class accuracies of the models as the compression increases are comparatively very less and for some compression rates the accuracies even gets an increase in values. Whereas, for the class Graphics in figure \ref{fig_compression_graphics_mcevit}, there is negligible drop in class accuracies of the models, it has almost stable accuracies as the compression increases. Altogether, analyzing the class accuracies shows that the rate of decrease in class accuracies with the increase in compression differs for different classes. And hence, how we can exploit this vulnerability of the decrease in accuracy with an increase in compression, which differs for each class, is the motivation for our proposed approach of a robust classification model for the forensic task of distinguishing natural and computer generated images. 

\subsection{Network architecture}
\label{sec_architecture_mcevit}

\begin{figure*}[!h]
\centering
\includegraphics[width=\linewidth]{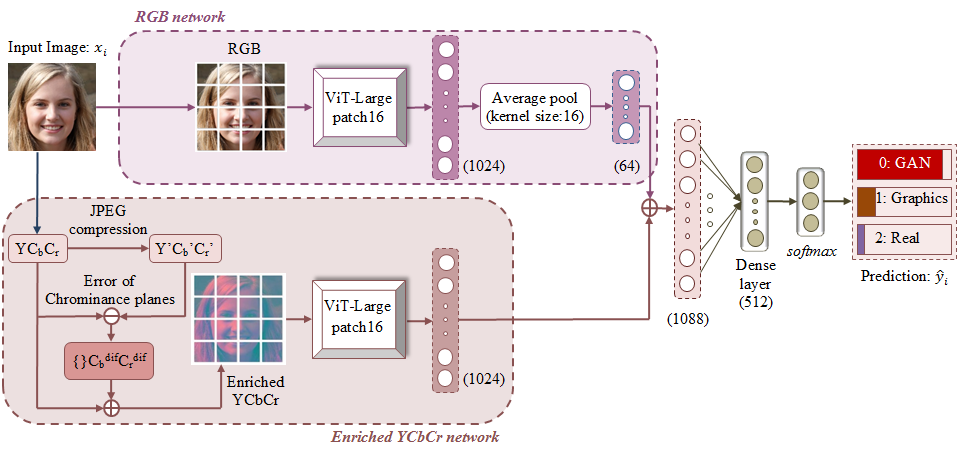}
\caption{ The overall architecture of \hl{the proposed model} Multi-Colorspace fused and Enriched Vision Transformer (\textit{MCE-ViT})}
\label{figure_architecture_mcevit}
\end{figure*}

We devise our methodology in such a way that the proposed model should obtain high classification accuracy and should also be highly robust. We propose a Multi-Colorspace fused and Enriched Vision Transformer (\textit{MCE-ViT}) model by parallely combining two transformer based networks that operates in two different color spaces, one for obtaining high classification accuracy and the other network dedicated to improve the robustness of classification. The entire architecture of our proposed model is shown in figure \ref{figure_architecture_mcevit}. For obtaining high classification accuracies we choose as the first network of the fused model, one of the recent state-of-the-art transformer based models ViT \cite{dosovitskiy2021image}. This network takes as input the RGB images, and we name this first ViT network as \textit{RGB ViT network}. We follow a transfer learning strategy where we choose the ViT network pre-trained on the Imagenet 21-k dataset \cite{ridnik2021imagenet} and fine-tune the network using the task specific GAN, Graphics and Real images dataset (detailed in section \ref{sec_dataset_mcevit}). Even though this first \textit{RGB ViT network} is found to obtain high classification accuracies for uncompressed images, it also has a performance decay when the images are JPEG compressed, as already seen in figure \ref{fig_compression_mcevit}. Since the rate of performance decay is different for different classes (as detailed in section \ref{sec_motivation_mcevit}), allowing the model to also learn about these differences in performance decay among the classes, would help to better identify these classes in compressed scenarios. Therefore we design a strategy for the second network in our fused model in such a way as to make the model learn these differences and thereby improve the robustness. 

\hl{As the first \textit{RGB ViT network} of the proposed fused model primarily focuses on high classification accuracies, the second network of the fused model is designed primarily to focus on the task of improving the robustness of the entire classification. Hence our strategy for the second network of the fused model is motivated by the fact that, since the rate of differences in accuracies between original and compressed versions of images are different for different classes, enriching the images in each class with this information on their corresponding rate of differences would be highly beneficial to distinguish natural and generated images with high classification accuracies, even for images in post-processed scenarios thereby improving robustness of the model. Accordingly, in the second network of our fused model, the input RGB images from the dataset are initially converted to the YCbCr color space. The motivation for choosing YCbCr color space is that it is the color space commonly used in the process of JPEG compression \mbox{\cite{aucklandJPEGJoint}}. The process of JPEG compression mainly downsamples the chrominance planes of this colorspace, i.e., Cb and Cr, because the changes brought up in these planes are less visually discernible \mbox{\cite{aucklandJPEGJoint,baeldungJPEGCompression}}. Hence enriching the network with the error of chrominance planes between the original images and their corresponding JPEG compressed versions, would enable the network to learn the distinguishable feature of the rate of differences in classification accuracies based on compression that differs for different classes. On this basis, our second \textit{Enriched YCbCr ViT network} enriches the YCbCr images with the information on the error of chrominance planes between the original images and their corresponding JPEG compressed versions as given in the following equation, and this second pre-trained ViT network is fine-tuned using these enriched YCbCr images.}

\begin{align}
    & \text{YCbCr}  \xrightarrow{\text{JPEG compression}} \text{Y'Cb'Cr'} \nonumber \\
    & \text{YCbCr} - \text{Y'Cb'Cr'} \xrightarrow{\text{Error of chrominance planes}} \text{\{\}Cb\textsuperscript{dif}Cr\textsuperscript{dif}} \nonumber \\
    & \text{YCbCr} + \text{\{\}Cb\textsuperscript{dif}Cr\textsuperscript{dif}} \xrightarrow{\text{Enrichment}} {\text{YCbCr}}_\text{enriched}
\end{align}

Both the first \textit{RGB ViT network} and the second \textit{Enriched YCbCr ViT network} produce an output feature vector. Even though the feature vector from the first RGB ViT network is very powerful to classify gan, graphics, and real images with high accuracy in uncompressed scenarios, in order to decrease its effect towards performance decay in compressed scenarios, we perform average pooling on this feature vector to obtain only the representative feature of the RGB ViT network. Thus, pooling the feature vector from the RGB ViT network helps to not affect the robustness of the model in compressed scenarios, as well as to maintain high model accuracies. Later, to build the fused model, the feature vectors from both the ViT networks are concatenated to form a single feature vector which is passed to a fully connected neural network with a dense layer and a 3-class output layer that determines the class label of the images. 

\hl{Our entire methodology including the conversion to YCbCr color space and enriching the YCbCr images with the error of chrominance planes between the original and their corresponding JPEG compressed versions can be applied irrespective of any image formats such as .png, .jpeg, etc. For example, for a .png image format, the first \textit{RGB ViT network} takes as input the RGB image as such. For the second \textit{Enriched YCbCr ViT network}, the .png image would be converted from RGB to YCbCr, which is later enriched with the error of chrominance planes of this YCbCr image and its corresponding JPEG compressed version. Furthermore, our novel methodology can also be applied to compressed input images as well (the major aim of this work being to develop a robust model towards post-processed images). This is because our methodology focuses on the rate of difference in accuracies as compression increases, where we could observe from the figure \mbox{\ref{fig_compression_class_mcevit}} that these rates are different for different classes. This rate of difference not only occurs between an original/uncompressed version of an image and its corresponding compressed version with a compression quality factor of 90, but also between compressed images with quality factors of 90 and 80, 80 and 70, etc. Accordingly, when a compressed image is presented to our proposed model, the second \textit{Enriched YCbCr ViT network} would perform its action of conversion to YCbCr, again compressing it, and enriching the YCbCr with the error of chrominance planes of these two. That is, since our methodology is centered on the rate of difference in classification accuracies as compression increases which differs for different classes, our methodology can handle uncompressed or compressed input images of different formats, thereby making the proposed model highly robust.}

\section{Experimental Settings}
\label{sec_exp_mcevit}

The proposed study utilizes the vision transformer `vit-large' with patch size 16 and input image size 224 x 224, for both \textit{RGB} and the \textit{Enriched YCbCr} networks of the fused model. The \textit{Enriched YCbCr ViT network} utilizes JPEG compression with a quality factor of 90. The study follows transfer learning strategy where both the ‘vit-large’ networks are pre-trained on ImageNet-21k \cite{ridnik2021imagenet} and fine-tuned on the task specific dataset used in the study. Each ViT network produces an output feature vector of size 1024. After manually analyzing the results of average pooling of the feature vector from the first \textit{RGB ViT network} with various kernel sizes and strides, we use a kernel size of 16 \hl{and a stride of 16,} as a representative setting, and the resultant feature vector of size 64 is concatenated with the feature vector from the second \textit{Enriched YCbCr ViT network}. The entire concatenated feature vector of size 1088 is fed to a dense layer of 512 neurons with \textit{ReLU} activation function, followed by an output dense layer of 3 neurons with \textit{softmax} activation function, making 559,107 number of trainable parameters. The other hyperparameters are batch size 16, \textit{categorical crossentropy} loss function, \textit{Adam} optimizer, and 50 epochs. The experiments are conducted on the deep learning workstation equipped with Intel Xeon Silver 4208 CPU at 2.10 GHz, 256 GB RAM, and two GPUs of NVIDIA Quadro RTX 5000 (16GB each), using the libraries Tensorflow (version 2.8.0), Keras (version 2.8.0), Torch (version 1.13.1+cu116), PyTorch Lightning (version 1.9.0), Transformer (version 4.17.0), and Albumentations (version 1.3.1).

\hl{The model performance analysis experiments in this work focus on analyzing the performance of the proposed model based on classification accuracy and comparing it with the baselines. The classification accuracy\footnote{\url{https://scikit-learn.org/stable/modules/model_evaluation.html\#accuracy-score}} of a multi-class (3-class in this work) classifier model is computed as follows:}
\begin{align} 
\textit{\text{Acc}} = \frac{1}{N}\sum_{i=1}^N \begin{cases}1 & \hat{y_i}=y_i,\\0 & else.\end{cases}
\end{align}
\hl{where, $y_i$ and $\hat{y_i}$ indicate the original/ground-truth class and predicted class of an i\textsuperscript{th} image respectively, and $N$ indicates the total number of images.}
  
\subsection{Baselines}
\label{sec_baselines_mcevit}

\hl{For the purpose of model performance comparison with baselines, we consider state-of-the-art works in the literature that classify natural and generated images. To the best of our knowledge, there is only a single work \mbox{\cite{gangan2022distinguishing}} classifying GAN, graphics, and real images and therefore, we choose this work for our baseline comparison experiments. We also perform model performance comparison with works that classify natural and either of the computer generated images (i.e., GAN or graphics only) \mbox{\cite{quan2018distinguishing,nataraj2019detecting,de2018exposing}}, and an off-the-shelf deep neural network architecture \mbox{\cite{szegedy2017inception}}. We detail these baselines below.}
%
\begin{itemize}
    \item MC-EffNet: Multi-colorspace fused EfficientNet networks proposed in \cite{gangan2022distinguishing}. This baseline work proposes a fused model by parallely combining three EfficientNet-B0 networks that operate in three different color spaces RGB, LCH, and HSV. After converting the images into LCH and HSV color spaces, the values are rescaled into the range 0 to 255 in order to supply into the EfficientNet network. The pre-trained EfficientNet-B0 networks are fine-tuned using the GAN, Graphics, and Real images of the dataset used in \cite{gangan2022distinguishing}. The feature vectors from all three EfficientNet networks are then concatenated and supplied to an output dense layer with 3 neurons and a \textit{softmax} activation function. This forms the MC-EffNet-1 network. Whereas, the MC-EffNet-2 network includes an additional preprocessing layer of laplacian of gaussian operation, before performing the color conversions to LCH and HSV. For both the networks the hyperparameters are, \textit{categorical cross-entropy} as loss function, \textit{Adam} optimizer with learning rate 0.001, 256 as batch size, and 100 epochs. 
    \item \hl{The other baselines chosen for this study include InceptionResNet \mbox{\cite{szegedy2017inception}}, Quan et al. \mbox{\cite{quan2018distinguishing}}, Nataraj et al. \mbox{\cite{nataraj2019detecting}} and Rezende et al. \mbox{\cite{de2018exposing}} that were chosen as the baselines in the work of  MC-EffNet \mbox{\cite{gangan2022distinguishing}}. InceptionResNet \mbox{\cite{szegedy2017inception}} is one of the high-accuracy models in the ImageNet classification task. We follow a transfer learning approach over the InceptionResNet-V2 architecture where the final output layer of 1000 neurons is replaced with 3 neurons to suit this 3-class classification task. The state-of-the-art works \mbox{\cite{quan2018distinguishing,nataraj2019detecting,de2018exposing}} in the literature classify natural images and either of the types of computer generated images (i.e., either only GAN images or only computer graphics images). Quan et al. \mbox{\cite{quan2018distinguishing}} propose a CNN based model to classify natural and computer graphics images. They predict results through local-to-global strategy where classification results of local patches are computed initially, and later the global prediction results of the whole image is acquired via majority voting. Nataraj et al. \mbox{\cite{nataraj2019detecting}} proposes a CNN based model to classify natural and GAN images by incorporating a co-occurrence feature of the RGB colorspace. Rezende et al. \mbox{\cite{de2018exposing}} classify natural and graphics images using a transfer learning approach on the ResNet-50 model in combination with an SVM classifier. All these baselines \mbox{\cite{quan2018distinguishing,nataraj2019detecting,de2018exposing}} are observed obtaining accuracy gains in their paper over many other related works in the literature. The experimental settings of all the baselines are followed in the same way as mentioned in the work of  MC-EffNet\footnote{\url{https://github.com/manjaryp/GANvsGraphicsvsReal/tree/main/Baselines}} \mbox{\cite{gangan2022distinguishing}}. Since all these baselines are not originally designed as 3-class classification tasks, for performance comparison experiments, the top dense layer of all these baselines is replaced using a final dense layer of 3 neurons with \textit{softmax} activation function, to conform to the 3-class classification task. Later, we fine-tune all the baselines with the dataset used in this study. This provides a general setting to analyze and compare the potential of the models in detecting generated images including both GAN and graphics.}
\end{itemize}

\section{Results and Discussions}
\label{sec_results_mcevit}

\begin{figure}[!h]
\centering
\subfloat[Confusion matrix]{\includegraphics[width=.48\textwidth,keepaspectratio]{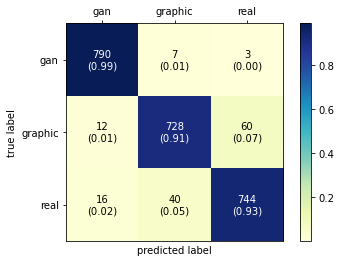}}
\hfil
\subfloat[DET curve]{\includegraphics[width=.5\textwidth,keepaspectratio]{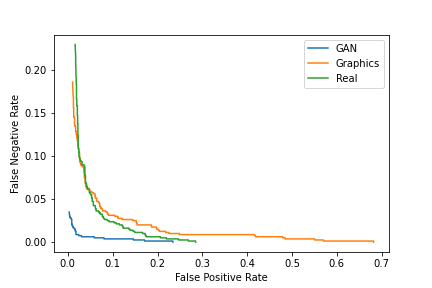}}
\caption{Confusion matrix and DET curve of \hl{the proposed model} \textit{MCE-ViT}}
\label{fig_confmat_mcevit}
\end{figure}

Our proposed model achieves a test accuracy of 94.25 percent. The 
confusion matrix \hl{and detection error trade-off (DET) curve} of the test results of our proposed model is shown in figure \ref{fig_confmat_mcevit}. Table \ref{table_baseline_comp_mcevit} presents a comparison of the test results of our model and the baselines in terms of total accuracy and class-wise accuracy. The test results indicate that our proposed model outperforms the baselines in terms of overall test accuracy and even in the case of individual class-wise accuracy. Our model obtains a gain of 4.87 percentage points in terms of overall accuracy when compared to MC-EffNet-2, the best performing model among the baselines. 

\begin{table}[!h]
\centering
\caption{\hl{Model performance accuracies of the proposed model and the baselines in percentage} (The highest accuracy is given in boldface)}
\label{table_baseline_comp_mcevit}
\begin{tabular}{lcccc}
\toprule
\multicolumn{1}{c}{Model} & \multicolumn{1}{c}{GAN} & \multicolumn{1}{c}{Graphics} & \multicolumn{1}{c}{Real} & \multicolumn{1}{c}{Total accuracy} \\
\midrule
\textit{MCE-ViT }(\hl{Proposed} model)      & \textbf{98.75} & \textbf{91.00} & \textbf{93.00} & \textbf{94.25} \\
MC-EffNet2 \cite{gangan2022distinguishing}       & 96.75          & 83.75          & 87.63          & 89.38          \\
MC-EffNet1 \cite{gangan2022distinguishing}       & 96.25          & 81.75          & 85.88          & 87.96          \\
InceptionResNet \cite{szegedy2017inception}      & 85.25          & 73.50          & 79.63          & 79.46          \\
Quan et al. \cite{quan2018distinguishing}        & 56.08          & 72.88          & 69.79          & 66.25          \\
Nataraj et al. \cite{nataraj2019detecting}       & 76.00          & 48.25          & 57.13          & 60.46          \\
Rezende et al. \cite{de2018exposing}             & 59.40          & 52.83          & 57.40          & 56.54          \\
\botrule
\end{tabular}
\end{table}

Among the three classes, GAN obtains the highest individual class accuracy of 98.75 percent, achieving a gain of 2 percentage points when compared to the highest class accuracy of 96.75 percent obtained by MC-EffNet-2 amongst the baselines. Class Graphics achieves an accuracy of 91.00 percent, a very high gain of 7.25 percentage points when compared to the highest class accuracy of 83.75 percent for the MC-EffNet-2 model amongst the baselines. Class Real achieves 93.00 percentage accuracy, i.e., a gain of 5.37 percentage points when compared to the highest class accuracy of 87.63 percent obtained by MC-EffNet-2 amongst the baselines.

\subsection{Robustness against Post-processing}
\label{sec_robust_mcevit}

Besides achieving high classification accuracies on original images or images that are not being post-processed, an efficient image forensic algorithm should also provide robust classification output over post-processed images. Hence, the robustness of the proposed model \textit{MCE-ViT} is evaluated towards the typical post-processing operation of JPEG compression and is compared against the baselines. \hl{Apart from the baseline models discussed in section \mbox{\ref{sec_baselines_mcevit}}, we also analyze the robustness of the proposed model over the off-the-shelf pre-trained vision transformer networks, ViT-B/16 and ViT-L/16 that are fine-tuned using the dataset used in this study. This is because we have observed in section \mbox{\ref{sec_motivation_mcevit}} that despite improved accuracies achievable by ViT they are prone to decline in performance with an increase in compression. So here, we also inspect whether our novel methodology of exploiting the vulnerability of the decrease in classification accuracies of image classes with the increase in compression, has helped to improve robustness. Here, the results of ViT-L/16 will serve as an ablation study for our methodology that incorporates colorspace fusion and enrichment of YCbCr with the error of chrominance planes information using ViT-L/16.}

\begin{figure}[!h]
\centering
\includegraphics[width=0.69\linewidth]{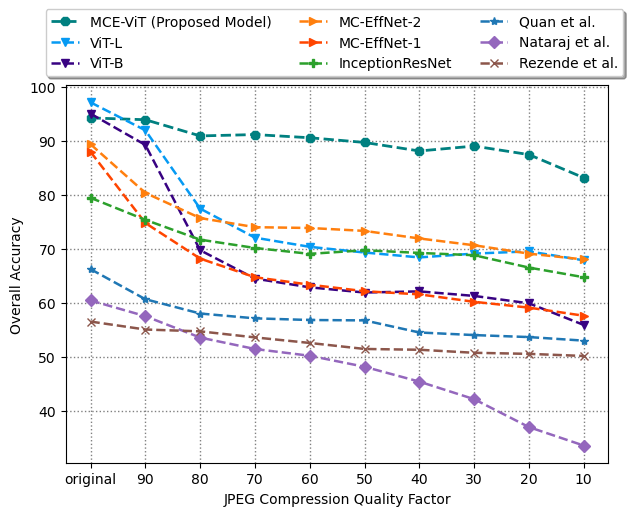}
\caption{Classification accuracies of \hl{the proposed} model and the baselines for various JPEG compression quality factors}
\label{fig_robustness_mcevit}
\end{figure}

Every model trained on uncompressed data (or original train data in the dataset) is tested separately over ten different test data created using various JPEG compression quality factors within the range 100 to 10, in steps of 10. Figure \ref{fig_robustness_mcevit} shows the robustness test results, where it can be observed that compared to the baselines, the proposed model achieves improved robustness towards post-processing based on JPEG compression, for all the compression quality factors. Here also, similar to the classification results of original uncompressed images (table \ref{table_baseline_comp_mcevit}), MC-EffNet-2 is the baseline model that shows the next improved robustness among all the baselines for different compression quality factors. \hl{When looking at the results of uncompressed images the accuracy of the proposed model is less than the ViT-L (ablation) and the ViT-B networks. We provide in table \mbox{\ref{table_robust_mcevit}}, the accuracies of our model and the ViT-L and ViT-B networks, for a better comparison. From the table, we can observe that for uncompressed images the proposed model has accuracy less than both ViT-L and ViT-B networks. This could be because, the feature vector from the ViT-L network even though is very powerful in classifying gan, graphics, and real images with high accuracy in uncompressed scenarios, since they fail in compressed scenarios our model has only taken a representative feature of this network through average pooling so that it provides better accuracy and also decrease its effect towards performance decay in compressed scenarios. But even then, our model has very significant robustness starting from a very low compression (quality factor 90) to a very high compression (quality factor 10). That is, even being built with a backbone of the ViT network, our methodology of exploiting the vulnerability of accuracy decline in image classes with an increase in compression, plays a vital role in providing significant robustness.}

\begin{table}[!h]
\centering
\caption{Accuracies over uncompressed/original images and compressed images (The highest accuracy is given in boldface)}
\label{table_robust_mcevit}
\setlength{\tabcolsep}{2.8pt}
\begin{tabular}{llllllllllll}
\toprule
\multirow{2}{*}{Model} 
    & \multirow{2}{*}{\begin{tabular}[c]{@{}c@{}}Original\\ images\end{tabular}} 
    & \multicolumn{9}{c}{Compressed images with quality factor:} \\
        &  
        & \multicolumn{1}{c}{90} 
        & \multicolumn{1}{c}{80} 
        & \multicolumn{1}{c}{70} 
        & \multicolumn{1}{c}{60} 
        & \multicolumn{1}{c}{50} 
        & \multicolumn{1}{c}{40} 
        & \multicolumn{1}{c}{30} 
        & \multicolumn{1}{c}{20}
        & \multicolumn{1}{c}{10} \\
\midrule
\textit{MCE-ViT} {\footnotesize(proposed)} & 94.25 & \textbf{93.92} & \textbf{90.92} & \textbf{91.17} & \textbf{90.58} & \textbf{89.71} & \textbf{88.13} & \textbf{89.04} & \textbf{87.46} & \textbf{83.17} \\
ViT-L & \textbf{97.13} & 91.96 & 77.46 & 72.08 & 70.38 & 69.33 & 68.42 & 69.13 & 69.54 & 67.88 \\
ViT-B & 95.04 & 89.25 & 69.75 & 64.50 & 62.88 & 61.92 & 62.13 & 61.29 & 59.92 & 55.92 \\
\botrule
\end{tabular}
\end{table}

\begin{figure}[!h]
\centering
\includegraphics[width=0.69\linewidth]{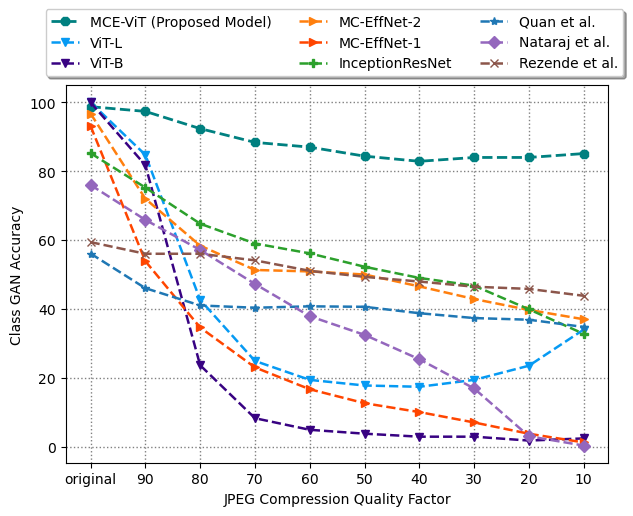}
\caption{GAN class accuracies of \hl{the proposed} model and the baselines for various JPEG compression quality factors}
\label{fig_robustness_gan_mcevit}
\end{figure}

Besides analyzing robustness against JPEG compression in terms of overall accuracy, we also analyze the robustness of our proposed model specifically for the class GAN, since it is the class that is most affected by post-processing operations and also the major contributing factor towards the decrease in overall accuracy, as already discussed in section \ref{sec_motivation_mcevit}. Figure \ref{fig_robustness_gan_mcevit} shows the robustness test results of our proposed model and the baselines, specifically for the class GAN. We can observe that our proposed model achieves very high class accuracy for all the compression factors. The class accuracy obtained for our model while classifying original images (no compression) is 98.75 percent. At the same time, for the images compressed at a compression factor of 10 (very high compression), the class accuracy of our model is 85.13 percent. That is, for a very high compression factor of 10, our model obtains a very high gain of 41.32 percentage points when compared to the next higher accuracy of 43.81 percent at compression factor 10 for the baseline Rezende et al. \cite{de2018exposing}. 

The class accuracy of the baseline ViT-L network (ablation) at a compression factor 10 is only 34.00 percent. That is, our proposed model even being built using the ViT-L network, but our methodology of color space fusion and the enrichment of YCbCr with the error of chrominance planes information between original and compressed images has proven to be highly advantageous in improving the robustness of the proposed model thereby achieving a class accuracy of 85.13 percent even at a compression factor of 10, i.e., a very huge gain of 51.13 percentage points when compared to the ViT-L network (ablation).

\begin{figure}[!h]
\centering
\includegraphics[width=0.68\linewidth]{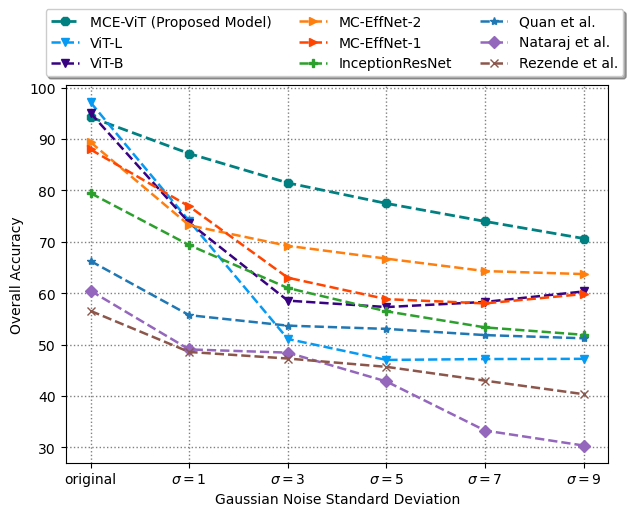}
\caption{Classification accuracies of \hl{the proposed} model and the baselines for various Gaussian noise standard deviations ($\sigma$)}
\label{fig_gausnois_mcevit}
\end{figure}

\hl{We also find the robustness of our model against another post-processing operation of addition of Gaussian noise and compare it with the baselines.} The accuracy of our proposed model and the baselines at various standard deviation ($\sigma$) values of Gaussian noise is shown in figure \ref{fig_gausnois_mcevit}. We can observe that compared to the baselines our model achieves better robustness against Gaussian noising also. \hl{Thus, this shows that, even if the images are post-processed with the compression or addition of gaussian noise, our novel methodology of exploiting the rate of difference in accuracies as compression increases, is powerful enough to distinguish between GAN, graphics, and real images.}

\subsection{Generalizability}
\label{sec_generalizability_mcevit}

This study also analyzes the generalizability of our proposed model \textit{MCE-ViT} by testing over unseen data. The proposed model \textit{MCE-ViT} which is fine-tuned on the dataset used in this study is tested over three other combinations of unseen GAN, Graphics, Real images, i.e., PG\textsuperscript{2} versus PRCG versus PIM-Google \cite{ma2017pose,ng2005columbia},  PG\textsuperscript{2} versus PRCG versus PIM-Personal \cite{ma2017pose,ng2005columbia}, and Cycle GAN versus Raise versus Level-Design \cite{zhu2017unpaired,dang2015raise,piaskiewicz2017level} datasets, with 160 images in each of the class of the three datasets, as experimented in the baseline work \cite{gangan2022distinguishing}. The generalizability test results are shown in table \ref{table_generalizability_mcevit}, where it can be observed that compared to the baselines our proposed model is able to obtain higher test accuracies. The results hence prove that our model has better generalizability, which can even help towards the future challenges in generated image categories.

\begin{table}[!h]
\centering
\caption{\hl{Generalizability test results of the proposed model and the baselines over three datasets in percentage} (The highest accuracies are given in boldface)}
\label{table_generalizability_mcevit}
\begin{tabular}{lccc}
\toprule
\multicolumn{1}{c}{Model} &
  \multicolumn{1}{c}{\begin{tabular}[c]{@{}c@{}}PG\textsuperscript{2} $\times$  PRCG $\times$ \\ PIM-Google\end{tabular}} &
  \multicolumn{1}{c}{\begin{tabular}[c]{@{}c@{}}PG\textsuperscript{2} $\times$  PRCG $\times$ \\ PIM-Personal\end{tabular}} &
  \multicolumn{1}{c}{\begin{tabular}[c]{@{}c@{}}Cycle GAN $\times$  Raise $\times$ \\ Level-Design\end{tabular}} \\
\midrule
\textit{MCE-ViT }(\hl{Proposed} model)      & \textbf{87.84} & \textbf{90.31} & \textbf{91.75} \\
MC-EffNet-2 \cite{gangan2022distinguishing}      & 81.04          & 85.21          & 84.79          \\
InceptionResNet \cite{szegedy2017inception} & 62.08          & 67.16          & 71.74          \\
Quan et al. \cite{quan2018distinguishing}      & 54.22          & 56.27          & 60.01          \\
Rezende et al. \cite{de2018exposing}   & 51.25          & 51.21          & 50.92          \\
Nataraj et al. \cite{nataraj2019detecting}   & 49.17          & 53.63          & 48.75          \\
\botrule
\end{tabular}
\end{table}

\subsection{Feature Visualization}
\label{sec_viz_mcevit}

\begin{figure*}[!h]
\centering
\fbox{\subfloat[Raw image features]{\includegraphics[width=0.475\textwidth,height=4.75cm]{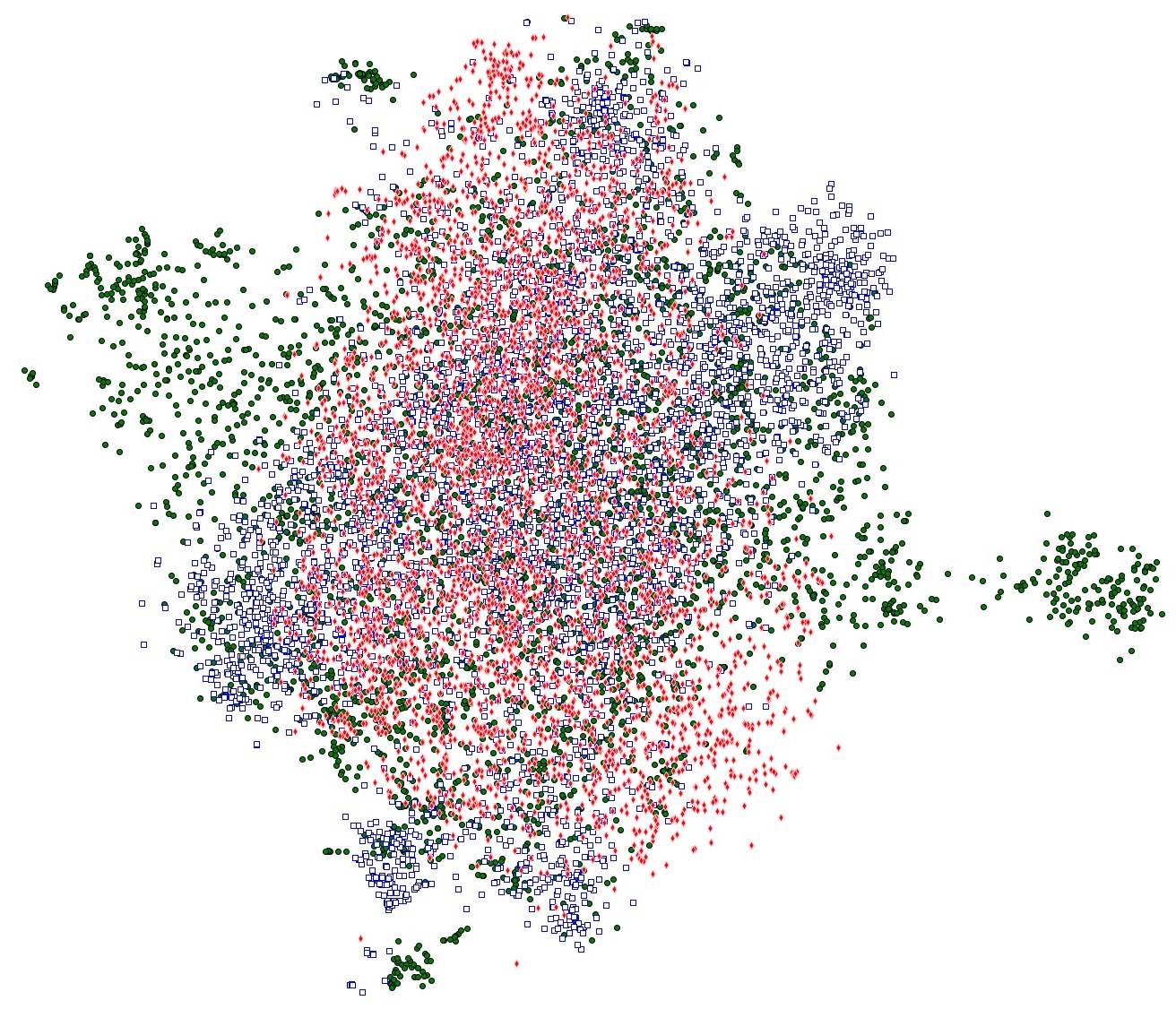}\label{fig_tsne_input_mcevit}}}
\hfil
\fbox{\subfloat[\textit{MC-EffNet-2} features \hl{(Baseline model)}]{\includegraphics[width=0.475\textwidth,height=4.75cm]{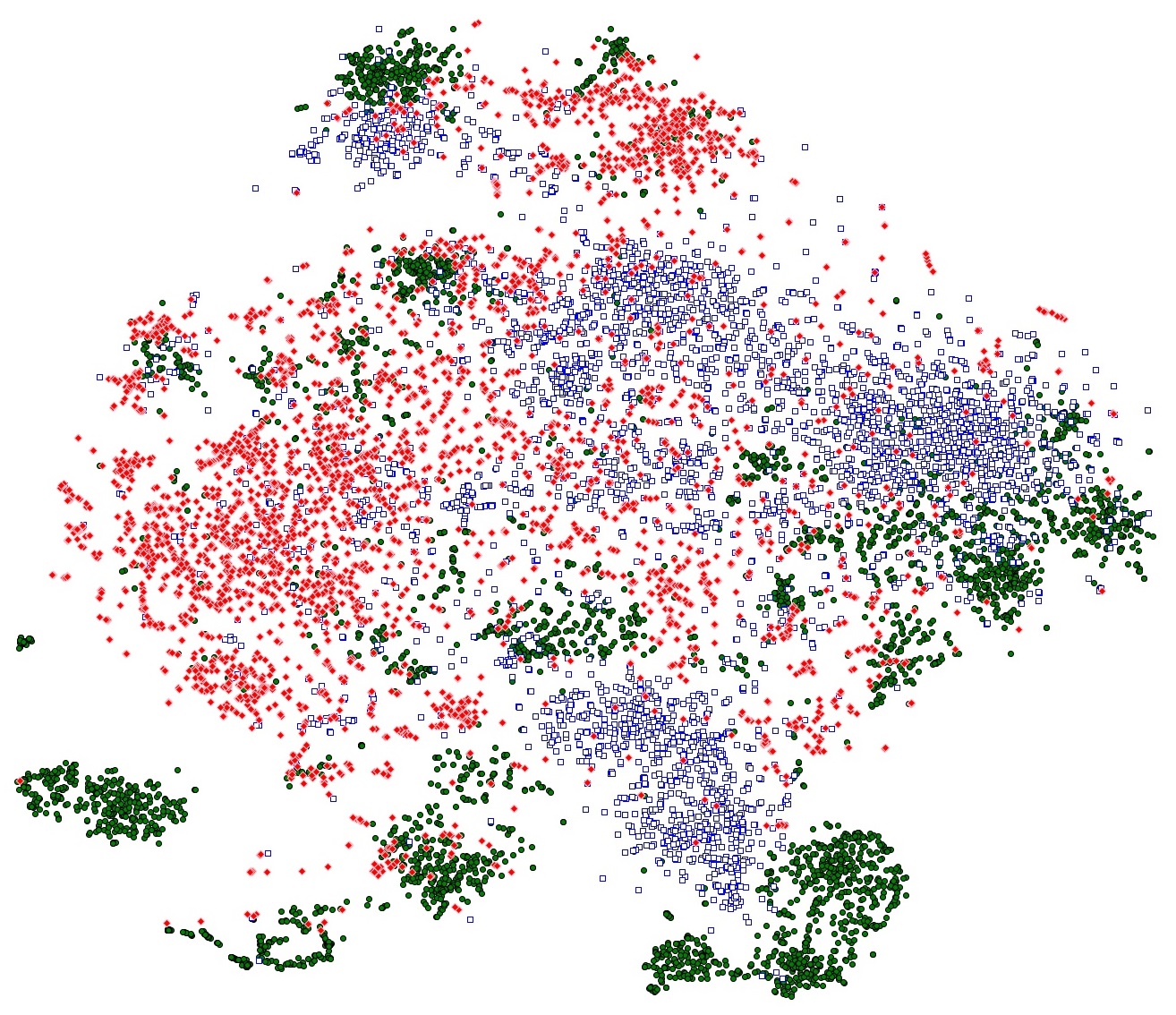}\label{fig_tsne_mceffnet2_mcevit}}}
\hfil
\fbox{\subfloat[\textit{MCE-ViT} features (\hl{Proposed} model)]{\includegraphics[width=0.475\textwidth,height=4.75cm]{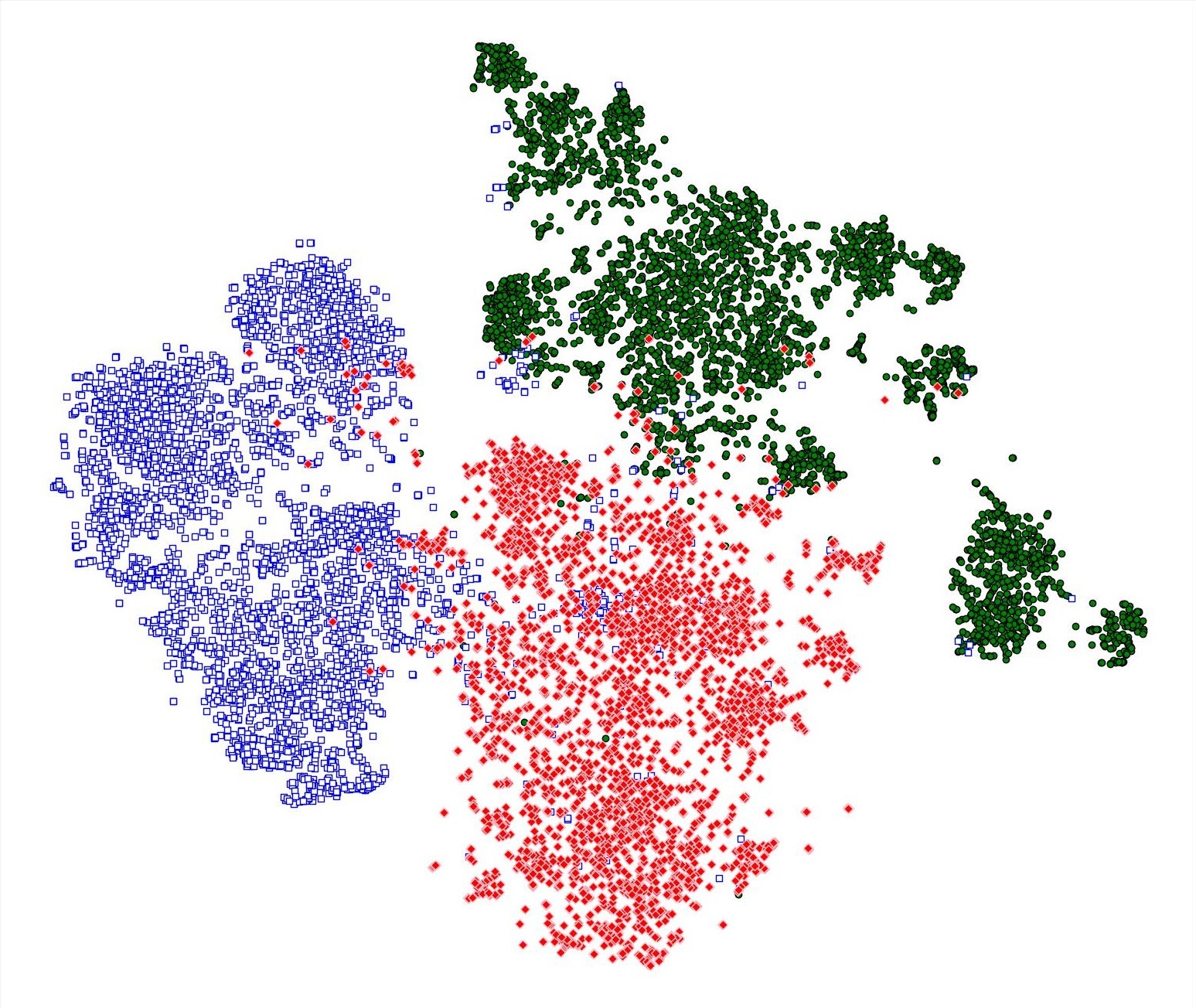}\label{fig_tsne_mcevit}}}
\caption{t-SNE visualizations of feature vectors. (\tikzsymbol{fill=cadmiumgreen} indicate GAN images, \tikzsymbol[rectangle]{minimum width=6pt,minimum height=6pt,blue,fill=white} indicate Graphics images and, \tikzsymbol[diamond]{pastelpink,fill=mediumcandyapplered,} indicates Real images)}
\label{fig_tsne_comp_mcevit}
\end{figure*}

The dimension of images that are input to the proposed \textit{MCE-ViT} model is 224 $\times$ 224 $\times$ 3, i.e., the length of the raw image features is 150528. \textit{MCE-ViT} projects these raw image features into a smaller dimension of length 1088, with an aim to provide better separability between the three classes. We analyze the separability potential of the feature vector of our model by comparing against the feature vectors of raw input image pixels and the best baseline model MC-EffNet-2 that obtains the next higher classification accuracy among all the baselines. The t-Distributed Stochastic Neighbor Embedding (t-SNE) \cite{van2008visualizing} dimensionality reduction technique can visualize high dimensional features into a two-dimensional plane and thereby helps to easily compare the separability potential of the feature vectors. Using the t-SNE technique we project the raw image features, feature vector output from the baseline model MC-EffNet-2 and feature vector output from our proposed model \textit{MCE-ViT}, into two-dimensional plots, shown in figure \ref{fig_tsne_comp_mcevit}. In each of the plots, three different colors are used to indicate three different classes, green circles are used to represent GAN image class, blue squares are used to represent Graphics images class, and pink diamonds are used to represent the class of Real images. From the t-SNE visualizations, it is clearly understandable that the raw image features do not have the potential to separate the classes, all the classes are clustered together, more particularly towards the center of the plot (figure \ref{fig_tsne_input_mcevit}). The output features projected from the baseline model MC-EffNet-2 (figure \ref{fig_tsne_mceffnet2_mcevit}) seem to produce separability between the classes better than the raw image pixels. The output features projected from our proposed model \textit{MCE-ViT} (figure \ref{fig_tsne_mcevit}) seem to produce much better separability between the three classes when compared to both raw pixels and the best baseline MC-EffNet-2. This demonstrates that our proposed model \textit{MCE-ViT} has better capability to transform the raw image pixels to a much better separable feature space for the forensic task of distinguishing natural images from computer-generated images including computer graphics and GAN images.

\subsection{Attention visualization}
\label{sec_attn_mcevit}

While visualizing the attention maps of both the networks of our fused model i.e., the \textit{RGB network} and the \textit{Enriched YCbCr network}, we could observe that compared to the \textit{RGB network}, the \textit{Enriched YCbCr network} could potentially identify more regions or information in an image that are helpful for classifying images into the classes GAN, Graphics or Real. Table \ref{table_attn_mcevit} shows a comparison of the attention map visualizations of both the \textit{RGB network} and the \textit{Enriched YCbCr network} for a set of GAN, Graphics and Real images. For example in the first generated image of a cat in the set of GAN images, the RGB network mainly captures the edges of the cat, whereas the \textit{Enriched YCbCr network} captures some of the information captured by the \textit{RGB network} and even more including from the \hl{background}, irrespective of the regions of the object in the image. Similarly, in the second generated face image, the forehead, right eye, and the image background are given more attention by the \textit{Enriched YCbCr network} than the \textit{RGB network}.

Similarly in the set of graphics images also, we can observe that the \textit{Enriched YCbCr network} is able to capture more relevant image information in detail than the \textit{RGB network}, such as the regions of uneven illuminations in the image. Also, mostly, the attention given to the captured image regions by the \textit{Enriched YCbCr network} are higher than the attention given by the \textit{RGB network}.

In the real class of images also we can observe that the \textit{Enriched YCbCr network} could capture more image information than the \textit{RGB network}. For example in the first image of the dog and the second image of the cow, the attention given to the image regions are higher for \textit{the Enriched YCbCr network} than the \textit{RGB network}. Also, more image regions relevant for classification are seen to be captured by the attention maps of the \textit{Enriched YCbCr network}.

Thus for all the classes we can observe that the image information captured by the \textit{Enriched YCbCr network} for this forensic task is comparatively much more relevant than the \textit{RGB network}. This, in fact, shows the powerful nature of our methodology incorporating the YCbCr color space transformation and enriching the color space with the information on the rate of differences between original and corresponding compressed versions of images in different categories, to capture image information that is highly relevant to the forensic task of classifying natural and generated images.

\setlength{\tabcolsep}{1pt}
\begin{table*}[!h]
\centering
\caption{Attention map visualizations of \textit{RGB network} and \textit{Enriched YCbCr network}}
\label{table_attn_mcevit}
\begin{tabular}{lll||lll}
\toprule
\multicolumn{1}{c}{\multirow{2}{*}{\begin{tabular}[c]{@{}c@{}}\footnotesize Original image\end{tabular}}} 
    & \multicolumn{2}{c||}{\footnotesize Attention map visualizations} 
    & \multicolumn{1}{c}{\multirow{2}{*}{\begin{tabular}[c]{@{}c@{}}\footnotesize Original image\end{tabular}}} 
    & \multicolumn{2}{c}{\footnotesize Attention map visualizations} 
    \\ \cmidrule{2-3} \cmidrule{5-6} 
\multicolumn{1}{c}{} 
    & \multicolumn{1}{c}{\textit{\footnotesize RGB network}} 
    & \multicolumn{1}{c||}{\begin{tabular}[c]{@{}c@{}}\textit{\footnotesize Enriched YCbCr}\\ \textit{\footnotesize network}\end{tabular}} 
    & \multicolumn{1}{c}{} 
    & \multicolumn{1}{c}{\textit{\footnotesize RGB network}} 
    & \multicolumn{1}{c}{\begin{tabular}[c]{@{}c@{}}\textit{\footnotesize Enriched YCbCr}\\ \textit{\footnotesize network}\end{tabular}} 
    \\ \midrule
\multicolumn{6}{c}{GAN} \\ \midrule
\includegraphics[width=22mm, height=22mm]{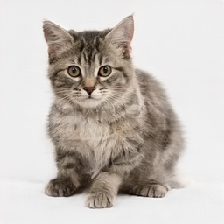} 
& \includegraphics[width=22mm, height=22mm]{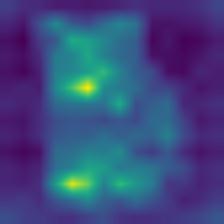}
& \includegraphics[width=22mm, height=22mm]{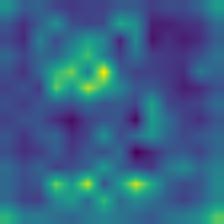}
& \includegraphics[width=22mm, height=22mm]{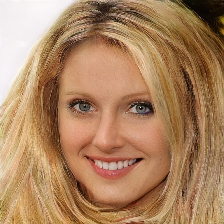}  
& \includegraphics[width=22mm, height=22mm]{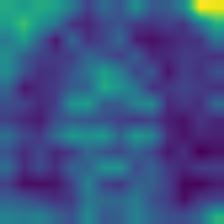} 
& \includegraphics[width=22mm, height=22mm]{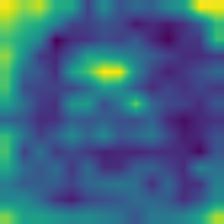}
\\ 
\includegraphics[width=22mm, height=22mm]{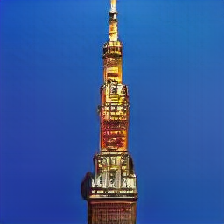}  
& \includegraphics[width=22mm, height=22mm]{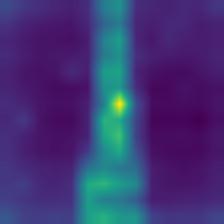} 
& \includegraphics[width=22mm, height=22mm]{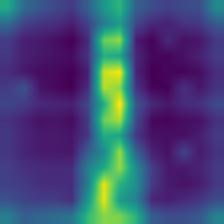}
& \includegraphics[width=22mm, height=22mm]{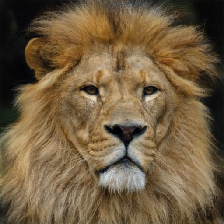}  
& \includegraphics[width=22mm, height=22mm]{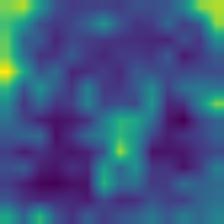} 
& \includegraphics[width=22mm, height=22mm]{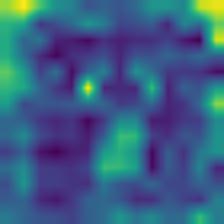}
\vspace{1pt} \\ \midrule
\multicolumn{6}{c}{Graphics} \\ \midrule
\includegraphics[width=22mm, height=22mm]{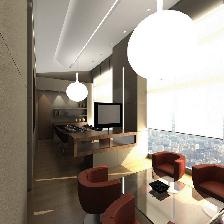}  
& \includegraphics[width=22mm, height=22mm]{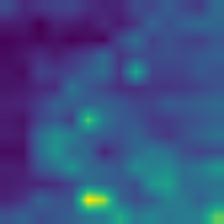} 
& \includegraphics[width=22mm, height=22mm]{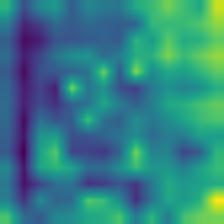}
& \includegraphics[width=22mm, height=22mm]{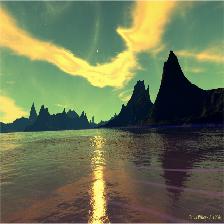}  
& \includegraphics[width=22mm, height=22mm]{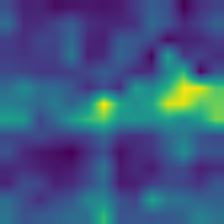} 
& \includegraphics[width=22mm, height=22mm]{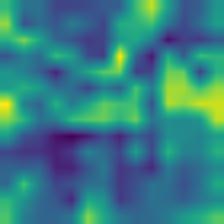}
\\ 
\includegraphics[width=22mm, height=22mm]{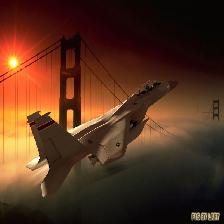}  
& \includegraphics[width=22mm, height=22mm]{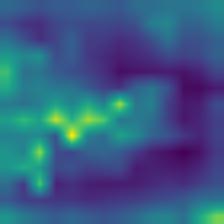} 
& \includegraphics[width=22mm, height=22mm]{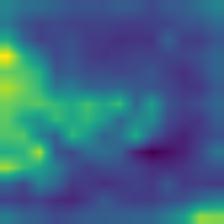}
& \includegraphics[width=22mm, height=22mm]{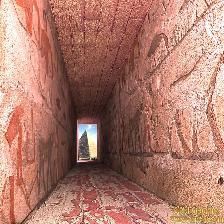}  
& \includegraphics[width=22mm, height=22mm]{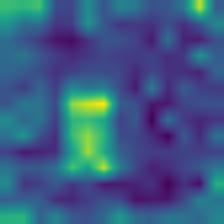} 
& \includegraphics[width=22mm, height=22mm]{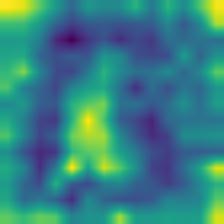}
\vspace{1pt} \\ \midrule
\multicolumn{6}{c}{Real} \\ \midrule
\includegraphics[width=22mm, height=22mm]{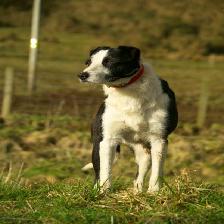}  
& \includegraphics[width=22mm, height=22mm]{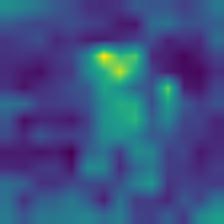} 
& \includegraphics[width=22mm, height=22mm]{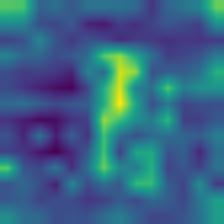}
& \includegraphics[width=22mm, height=22mm]{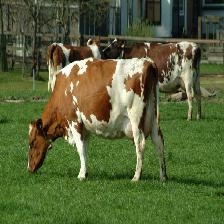}  
& \includegraphics[width=22mm, height=22mm]{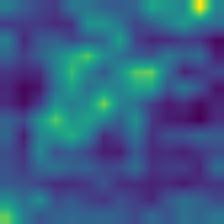} 
& \includegraphics[width=22mm, height=22mm]{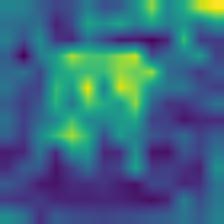}
\\ 
\includegraphics[width=22mm, height=22mm]{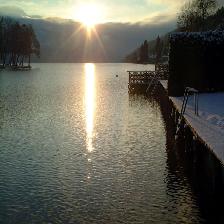}  
& \includegraphics[width=22mm, height=22mm]{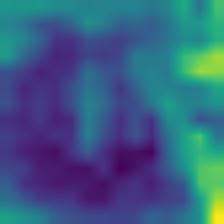} 
& \includegraphics[width=22mm, height=22mm]{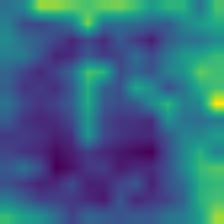}
& \includegraphics[width=22mm, height=22mm]{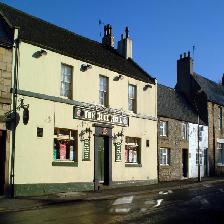}  
& \includegraphics[width=22mm, height=22mm]{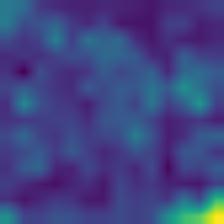} 
& \includegraphics[width=22mm, height=22mm]{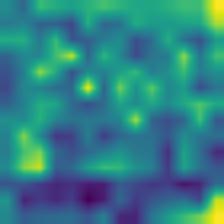}
\\ \botrule
\end{tabular}
\end{table*}

\section{Conclusion}
\label{sec_conclusion_mcevit}

In this work a robust approach towards distinguishing natural and computer generated images including both, computer graphics and GAN generated images is proposed, unlike the works in the literature to distinguish natural and computer generated images that consider only either of the generated images category. \hl{Our novel methodology focuses on exploiting the vulnerability of decrease in classification accuracies of image classes with the increase in compression factors, to majorly focus on building a forensic classifier system that is highly robust.} The proposed work utilized a fusion of two vision transformers one of them operating in RGB color space and the other in YCbCr color space. Transformer based architectures can provide good classification performance, while our methodology focuses on improving the robustness of the proposed model against post-processing operations such as JPEG compression by maintaining a high model accuracy because usually the forensic algorithms and tools even with high performance accuracies are fooled using these post-processed images. Experiments are conducted to analyze the performance of the model and are compared against a set of baselines. The proposed model achieves higher accuracy than the baselines and is found to be highly robust and generalizable. Visualizing the features showed better separability capability of the proposed model than the baseline. The work also studied the attention map visualizations of the networks of the fused model and observed that the proposed methodology could capture more image information relevant to the forensic task of classifying natural and generated images. To aid future research, the relevant materials of this study including the source code are made publicly available at \url{https://github.com/manjaryp/MCE-ViT} and \url{https://dcs.uoc.ac.in/cida/projects/dif/mcevit.html}. \hl{In the future, we plan to use and integrate our methodology to develop a web tool where users can test in real-time any image doubted for the authenticity of whether it is natural or computer generated. Additionally, we plan to explore other challenges in the area of digital image forensics, such as identifying recaptured images and computer generated videos.}








\subsection*{Acknowledgment} 
This work was supported by the Women Scientist Scheme-A (WOS-A) for Research in Basic/Applied Science from the Department of Science and Technology (DST) of the Government of India under the Grant SR/WOS-A/PM-62/2018.

\bibliography{references_mcevit}

\end{document}